\documentclass[conference]{IEEEtran}
\IEEEoverridecommandlockouts

\usepackage{cite}
\usepackage{amsmath,amssymb,amsfonts}
\usepackage{algorithmic}
\usepackage{graphicx}
\usepackage{textcomp}
\usepackage{url}
\usepackage{xcolor}
\usepackage[table]{xcolor}
\def\BibTeX{{\rm B\kern-.05em{\sc i\kern-.025em b}\kern-.08em
    T\kern-.1667em\lower.7ex\hbox{E}\kern-.125emX}}
\begin{document}

\title{Continuous Conditioning of VLAs with Augmenting EMG and Visual Task Descriptors}

\author{\IEEEauthorblockN{Edward W. Staley, Connor O. Pyles, Rahul Hingorani, Frank Camargo, Griffin Milsap, \\ Jared Markowitz, Matthew S. Fifer, and Michael Wolmetz}
\IEEEauthorblockA{\textit{Johns Hopkins Applied Physics Laboratory} \\
Laurel, MD, USA \\
first.last@jhuapl.edu}
}


\maketitle

\begin{abstract}
Vision-Language-Action (VLA) models rely strongly on language for describing task information, despite having multimodal inputs. We hypothesize that other modalities in the state space may present opportunities for supplemental task conditioning, which may be particularly relevant in cluttered or otherwise ambiguous scenes. We introduce two tuned models to test this hypothesis: (1) an electrophysiology-conditioned VLA (EC-VLA) that incorporates 8-channel electromyography envelopes as continuous conditioning input concatenated to the proprioceptive vector, and (2) a visually-annotated VLA (VA-VLA) that incorporates visual segmentation annotations to the image inputs. On a cube-selection task evaluated across three participants, EC-VLA matches a language-prompted baseline in uncluttered, in-distribution conditions and substantially outperforms it in cluttered, out-of-distribution scenes. Similarly, VA-VLA shows modest improvements over a language-prompted baseline in in-distribution scenes with substantial improvement in cluttered, out-of-distribution trials. Together, these results provide strong evidence for the potential benefit of task-conditioning beyond language.
\end{abstract}


\section{Introduction}

Vision-language-action (VLA) models have recently gained significant traction as a means for generalized robotic control, and can serve as a base model to be tuned for a downstream task \cite{rt1, openvla, octo, pi0, pi05, smolvla, molmoact}. Such tuned policies inherit their high-level state structure from the base model: a low-level proprioceptive vector, one or more camera views, and a task description given as language. The majority of VLA parameters are often initialized from a large language model (LLM), forming the core backbone of the model. At deployment time, language is used to define what the robot needs to accomplish in the environment and is typically given as an up-front command that stays fixed during task execution.

However, as these systems continue to become more capable and take on increasingly complex tasks, there will be times in which language is insufficient as a task-defining condition. Cluttered scenes and nuanced tasks might require detailed explanation that a human would more naturally express with a gesture, picture, or gaze \cite{m4bench, moo, gaze2act}.

\begin{figure}[htbp]
    \centering
    \includegraphics[width=0.5\textwidth]{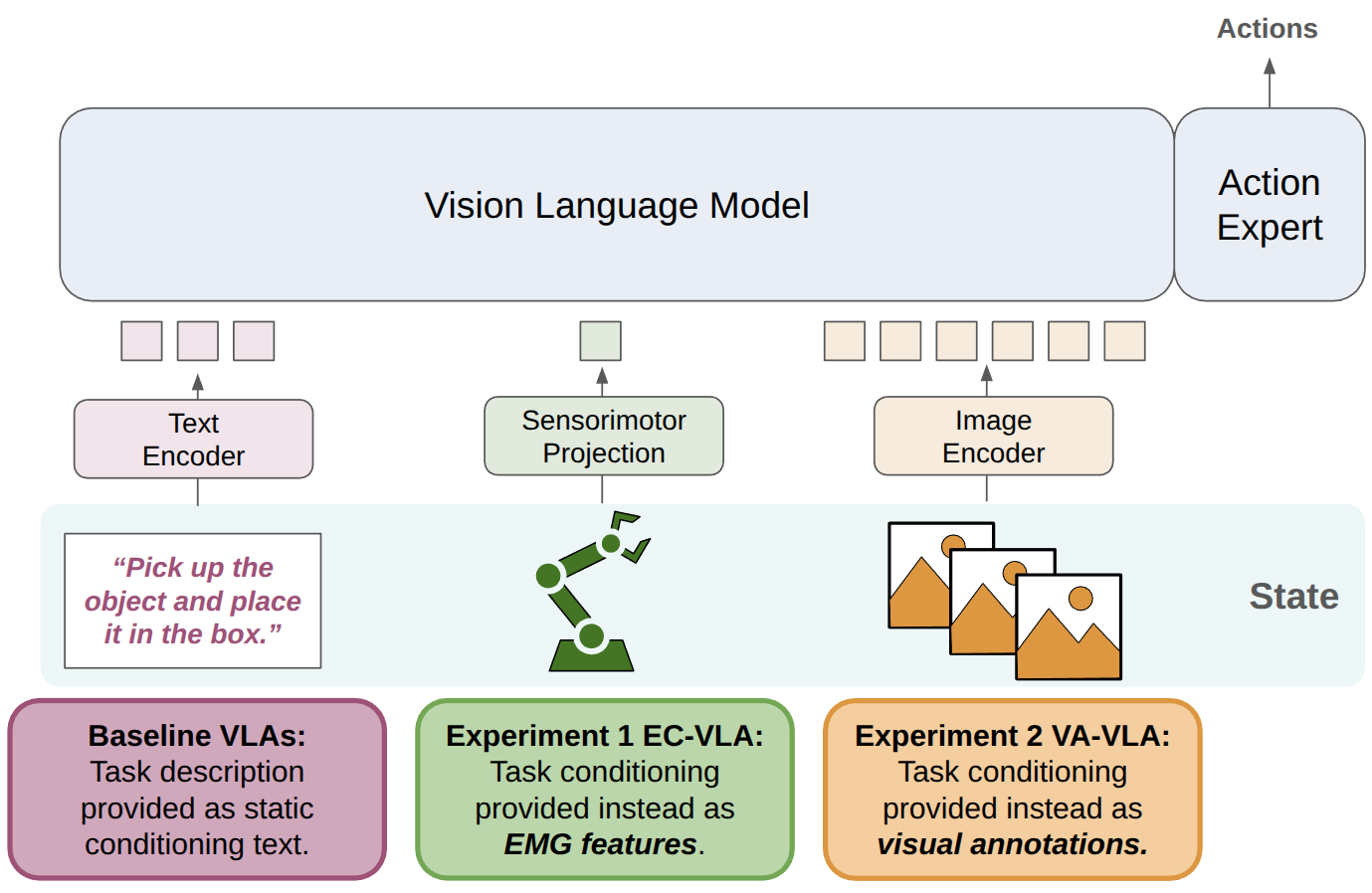}
    \caption{An overview of vision-language-action (VLA) model architectures and how we propose to modify the inclusion of task information.}
    \label{fig:summary}
\end{figure}

We explore tuning VLAs with tasks defined by continuous conditioning signals incorporated into state modalities beyond text. As shown in Figure \ref{fig:summary}, we construct two experiments: one with task conditioning in the proprioceptive vector and one with conditioning in the image modality, with only generic task information given in the text input. We compare these to tuned models that use the more standard descriptive text inputs. We find that these alternative modalities for task conditioning, supplied as continuous state information, generally yield stronger performance than the baseline tuned policies, especially in cluttered scenes.



\section{Background}

\subsection{Vision Language Action Models}

In our experiments we utilize SmolVLA \cite{smolvla} and $\pi_{0.5}$ \cite{pi05}, which are both action-expert style VLAs that use a flow-matching \cite{flowmatching} head on top of a VLM backbone. This architecture was first introduced in Octo \cite{octo} (using a diffusion policy head) and $\pi_0$ \cite{pi0}. It contrasts with earlier autoregressive techniques like OpenVLA \cite{openvla}, which output discrete tokens that need to be converted into actions afterwards.


\subsection{Electrophysiological Signals and Robot Policies}

Many prior works explore shared control between a robotic system and user data collected through electromyography (EMG), (electroencephalogram) EEG, or brain-computer interface (BCI) signals.  These inputs provide intent or feedback, often falling in the domain of assistive robotics. Xu et al. \cite{xu} use EEG signals to direct a robotic system towards grasping a target object. Akinola et al. \cite{akinola} use EEG as a way for a human to note an incorrect action, a signal which is then used to further improve the policy.

More recent and relevant to this work, Lee et al. \cite{lee} combine EEG with an AI copilot, in a similar setup to our first experiment. M4Bench \cite{m4bench} explored eye tracking and EMG as input modalities in a multirobot pick-and-place problem. The extension by Douglas et al. \cite{douglas} integrates EEG, EMG, and eye tracking in a system to control Franka arms in simulated kitchens with a focus on allocating control to either the human or policy. Our EC-VLA model is a robotic policy that directly receives EMG inputs as state information.

\subsection{Visual Annotations and Robot Policies}

Several prior works have examined visual prompting to describe tasks to a robotic policy. KITE \cite{kite} trains a policy that conditions on imagery and keypoints rather than language directly. MOKA \cite{moka} also annotates images with key points, asking a VLM to select among a set of key points to identify which are most useful. RT-Trajectory \cite{rttrajectory} uses 2D curves drawn over a scene that roughly depict how a robot should move. 

Works similar to our second experiment include FOCA \cite{foca}, which uses segmentation masks from RoboCasa at training time to supervise predicted features, and RoboGround \cite{roboground}, which uses segmentation-style masks to describe a target object and destination in RoboCasa. RoboGround conditions the policy on these masks and a language instruction, while in this work we explore omitting language when visual prompts are supplied. Also similar is MOO \cite{moo}, which directly describes a target object with a segmentation mask and omits this identity from language inputs to an RT-1\cite{rt1}-style policy. Gaze2Act \cite{gaze2act} includes a human in the loop who provides conditioning signals to help select an object while the text prompt is underspecified. Its human focus aligns well with our first experiment, but its signal is supplied in a manner closer to our second experiment: a human gaze is first translated into a visual mask and then supplied as an image.


\section{Methods}

We conduct experiments that specialize pretrained action-expert VLA models on tasks in which the task specification is provided via continuous conditioning by a modality that augments an underspecified language prompt. We compare these systems against the more standard paradigm of tuning with fully descriptive text prompts. 


Formally, we tune a model $VLA_\theta (A_t | O_t, k)$ which models the distribution of possible action chunks $A_t = [a_{t}, a_{t+1}, a_{t+2}, ... a_{t+H-1}]$ that consist of actions starting at the current timestep $t$ and continuing for some horizon (chunk size) $H$. These are conditioned on a current observation, $O_t$, which itself can be decomposed into three information sources: $O_t = (\ell_t, I_t, q_t)$, where $\ell_t$ is a language prompt describing the task, $I_t$ is one or more image observations, and $q_t$ is a proprioceptive vector input from the robotic system. $k \in [0,1]$ describes a timestep in the generative process used to generate $A_t$, which for all models in this work is a flow-matching procedure \cite{flowmatching}. 

To tune such a model, we collect a dataset of $N$ demonstrations $\mathcal{D} = \{(\ell,I,q,a)_{0:T}\}^N$, each consisting of tuples containing language, imagery, propriceptive state, and actions at each timestep $t$. The tuning objective is to minimize a flow matching loss which transports normally-distributed noise to the distribution of possible action chunks, which is defined with respect to $k$ as:

\begin{equation}
\mathcal{L}_{\mathrm{FM}}
=
\left\|
v_\theta(a_k, k, \ell,I,q)
-
(a_t-a_0)
\right\|_2^2
\end{equation}

where the prior $p_0$ is typically $\mathcal{N}(0,1)$, and the model outputs a velocity $v$.

In our two experiments, we examine replacing the descriptive text task description with a constant underspecified text prompt while augmenting another modality to be additionally informative. This results in a baseline dataset $\mathcal{D}$ and an augmented one $\mathcal{D}_{aug}$, such that $d \sim \mathcal{D}_{aug}=(\ell^u,I,q_{aug},a)_{0:T}$ or $d \sim \mathcal{D}_{aug}=(\ell^u,I_{aug},q,a)_{0:T}$ is a demonstration from the augmented dataset. Such a demonstrations has underspecified text $\ell^u$ and an augmented modality $q_{aug}$ (if the state vector is augmented) or $I_{aug}$ (if the imagery is augmented). In each experiment we tune a model separately on both dataset types and compare performance on the target tasks.

For both experiments, we also explore performance in cluttered scenes that are not represented in the training data. We hypothesize that, in cluttered environments, text may fall short of other modalities in its ability to specify a task. We grade all trials using a task-completion score that assigns partial credit to failures in four stages, adapted from MolmoAct \cite{molmoact}. Scoring and criteria for this partial credit scheme are described in Table \ref{tab:task_completion}.

\begin{table}[]
\centering
\caption{Task completion scoring system.}
\begin{tabular}{|c|c|c|c|}
\hline
\multicolumn{1}{|c|}{\textbf{Completion}} & \textbf{} & \textbf{Criteria}           & \textbf{Criteria} \\
\multicolumn{1}{|c|}{\textbf{Stage}} & \textbf{Score} & \textbf{(EC-VLA)}           & \textbf{(VA-VLA)} \\ \hline
3 & 1.0                   & Success                        & Success              \\ \hline
2 & 0.75                  & Touched target & Lifted target \\ \hline
1 & 0.5                   & Moved toward target    & Moved target           \\ \hline
0 & 0.0                   & Failure                        & Failure              \\ \hline
\end{tabular}

\label{tab:task_completion}
\end{table}

\subsection{Experiment 1: Electrophysiology-Conditioned VLA}

\subsubsection{EC-VLA Task}
In this experiment, we examine replacing the descriptive text task description with a constant underspecified text prompt and an 8-dimensional EMG signal added to the state information. An augmented demonstration consists of $d \sim \mathcal{D}_{EC}=(\ell^u,I,q_{EC},a)_{0:T}$, where $\ell^u$ denotes an underspecified constant text prompt and $q_{EC}=[q;EMG]$ is a concatenation of the standard proprioceptive vector and the EMG features. $EC$ is used instead of $aug$ to note the specific augmentations for experiment 1, concerning electrophysiological conditioning (EC). 

We apply the SO-101 arm to a block-grasping task in which four identical cubes are arranged along a linear workspace. The system is tasked with picking a given cube and placing it into a receptacle. A baseline descriptive prompt $\ell$ describes which cube to grasp, such as \textit{"Pick up the second to the left cube and place it in the box."}, while $\ell^u$ is simply the constant \textit{"Pick up the cube and place it in the box"}.

Our system runs at 15 Hz and includes two RGB camera views comprising $I_t$ (one overhead and one wrist-mounted). The actions to the robot $a_t$ are specified as 6-dimensional vectors describing desired positions for all six degrees of freedom. The baseline $q$ is a 6-dimensional proprioceptive vector that includes five joint angles and gripper status, and as described above, $q_{EC}$ contains eight additional inputs comprising EMG signals from a supervising human. The EMG input represents intent signals that the human can provide via wrist gestures, guiding the arm to correct left, correct right, or select a block. 

\begin{figure}[htbp]
    \centering
    \includegraphics[width=0.5\textwidth]{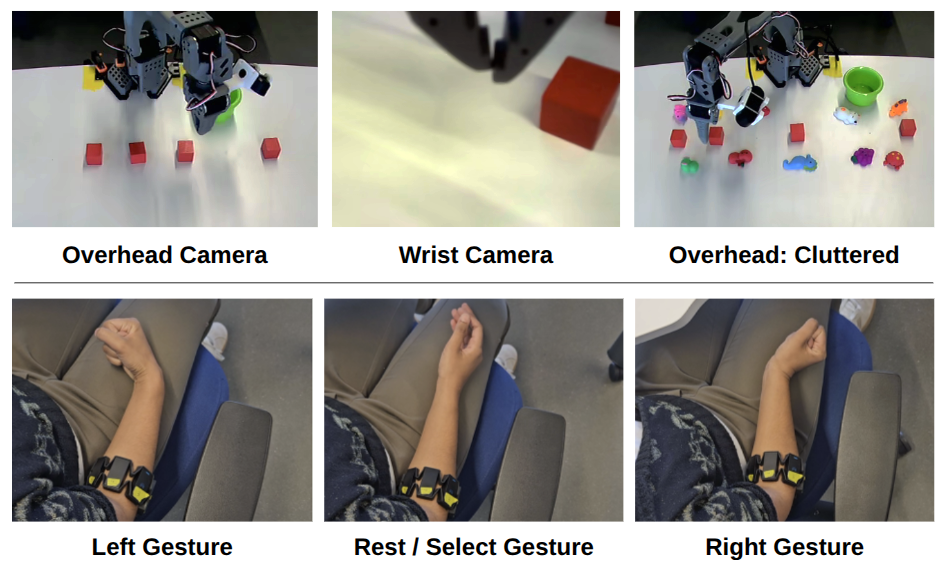}
    \caption{Camera views and gestures from the first experiment. Top row: Visual state information from the scene, showing an overhead view of four cubes in an uncluttered setting on the left, a typical wrist view in the center, and an example of a cluttered scene on the right. Bottom row: Wrist gestures used by participants to condition EC-VLA during training and test.}
    \label{fig:ec_detail}
\end{figure}

We collect three datasets $\mathcal{D}_{EC}^i$, one for each of three participants providing input signals. Details of this data collection and EMG featurization are provided in the Appendix (\ref{sec:ec_data}). We independently train SmolVLA on each of these datasets, as well as the unaugmented form of the first dataset as a baseline. We then evaluate the models on the target task, as well as on a cluttered variation. We refer to the models tuned on EMG data and generic text as an electrophysiology-conditioned VLA, or EC-VLA. Figure \ref{fig:ec_detail} shows examples of observation images, including a cluttered scene, as well as the range of gestures provided by a participant.

\subsubsection{EC-VLA Evaluation}

The above procedures result in four models: three participant-specific EC-VLA models and one baseline. We evaluate each on 30 trials, consisting of 20 in-distribution (uncluttered) trials and 10 cluttered trials in which additional objects have been added to the workspace. During evaluation inference, continuous EMG data is provided by the corresponding participant for each EC-VLA model. As described above, we collect metrics following a task completion score adapted from MolmoAct \cite{molmoact} in addition to success rate. See Table \ref{tab:task_completion}.




\subsection{Experiment 2: Visual-Annotation VLA}

\subsubsection{VA-VLA Tasks and Datasets}

In experiment two, we examine replacing the descriptive text task description with a constant underspecified text prompt and visual augmentations describing what object to pick up and where to place it: $d \sim \mathcal{D}_{VA}=(\ell^u,I_{VA},q,a)_{0:T}$, where $\ell^u$ denotes an underspecified constant text prompt and $I_{VA}$ uses visually augmented images highlighting the target object and destination. $VA$ is used instead of $aug$ to note the specific augmentations for experiment 2, concerning visual annotations (VA).


We utilize 16 pick-and-place tasks within the simulated RoboCasa environment. These tasks involve grasping a target object and placing it in a target location, a primitive step in many cooking tasks. 
A baseline descriptive prompt $\ell$ describes which object should be picked and where it should be placed, such as \textit{"Pick up the apple and place it on the plate on the counter."}, while $\ell^u$ is simply the constant \textit{"Pick the object and place it."}, regardless of task. For a complete list of the 16 tasks, please see the per-task performance results in the appendix.

These tasks are performed with a Franka Panda arm on a mobile base. The observations provided at 20 Hz include three rgb views comprising $I_t$ or ${I_t}_{VA}$ depending on if they include annotations and a 16-dimensional proprioceptive vector $q_t$. Robot actions $a_t$ are specified as 12-dimensional vectors. Episodes in train and evaluation use randomized layouts, interior styles, and object assets. RoboCasa provides approximately 100 human-teleoperated demonstrations per task, so our datasets $\mathcal{D}$ and $\mathcal{D}_{VA}$ are roughly 1600 example episodes.


We tune $\pi_{0.5}$ separately on $\mathcal{D}$ and $\mathcal{D}_{VA}$. To construct the visual annotations we modify the left and right images to highlight the target object in green and the location to place the object in blue. Examples of tasks and their original/annotated visuals are shown in the Appendix in \ref{fig:vaimagery}.


\subsubsection{VA-VLA Evaluation}

We evaluate both models on all 16 tasks individually for 50 randomized episodes (800 episodes in total). As in EC-VLA above, we gather partial success criteria as well as overall success rates (explained in Table \ref{tab:task_completion}). The baseline model is given descriptive text while the visual-annotation model is given underspecified text and altered visuals that highlight objects based on segmentation masks. 


We also run these same trials in eight modified tasks that include three additional random objects initialized near the target object. These “cluttered” tasks were selected because they have space available to spawn additional objects. Tasks that initialize a target object in a tight space (such as in a container or an appliance) were not tested in a cluttered configuration.


\section{Results}

\begin{figure}[htbp]
    \centering
    \includegraphics[width=0.5\textwidth]{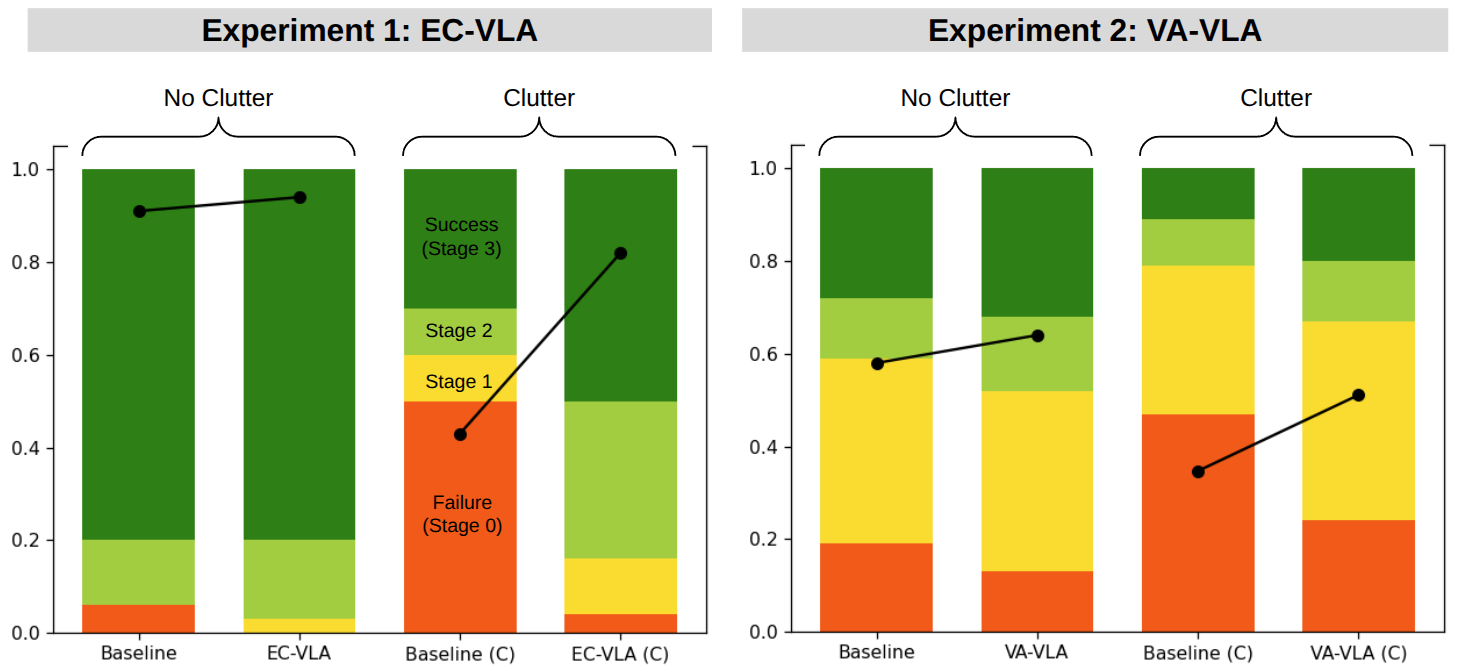}
    \caption{Behavior detail for the EC-VLA (left) and VA-VLA (right) trials, showing various partial success criteria as portions of trials. These include failure (red), stage 1 partial credit (yellow), stage 2 partial credit (light green), and success (dark green). The black lines show changes in task completion score, which is an aggregate score of these behaviors. EC-VLA values are averages of all three participant models, and VA-VLA values are only taken from the tasks that can add clutter. (C) denotes cluttered trials}
    \label{fig:bars}
\end{figure}

We find broadly that our continuous conditioning signals result in better performance compared to a descriptive text prompt. This increase is marginal in a typical setting (a few percent), but more substantial in cluttered scenes. A full breakdown of our results, including partial task completion values, is provided in Table \ref{tab:results}. We also provide a visual in Figure \ref{fig:bars} depicting the portion of trials that reach a particular stage in our scoring systems. 

\begin{table}[]
\centering
\footnotesize
\caption{Evaluation Results between tuned baselines and our augmented models.}
\hspace*{-0.2cm}
\begin{tabular}{|@{\hspace{2pt}}l@{\hspace{2pt}}|c|c|c|c|c|}

\hline
\rowcolor{black!10}\multicolumn{1}{|c|}{\textbf{Model}}                   & \textbf{Completion} & \textbf{Success} & \textbf{Stage 2} & \textbf{Stage 1} & \textbf{Failure} \\
\rowcolor{black!10}\multicolumn{1}{|c|}{\textbf{Evaluated}}                    & \textbf{Score} & \textbf{Rate} & \textbf{Rate} & \textbf{Rate} & \textbf{Rate} \\ \hline \hline
\rowcolor{black!10}\multicolumn{1}{|l|}{\textbf{\textit{Experiment 1}}} & & & & & \\ \hline \hline
SmolVLA       & 0.91             & 0.80          & 0.94         & 0.94         & 0.06         \\ \hline
EC-VLA                            & 0.94             & 0.80          & 0.97         & 1.0            & 0.00            \\ \hline
\multicolumn{1}{|r|}{\textit{\textcolor{blue}{Delta}}}           & \textit{\textcolor{blue}{+0.03}}   & \textit{\textcolor{blue}{0.00}}      & \textit{\textcolor{blue}{+0.03}}   & \textit{\textcolor{blue}{+0.06}}   & \textit{\textcolor{blue}{-0.06}}    \\ \hline \hline
SmolVLA (C)\footnotemark[1]              & 0.43             & 0.30          & 0.40          & 0.50          & 0.50          \\ \hline
EC-VLA (C)                & 0.82             & 0.50          & 0.84         & 0.96         & 0.04         \\ \hline
\multicolumn{1}{|r|}{\textit{\textcolor{blue}{Delta}}}    & \textit{\textcolor{blue}{+0.39}}        & \textit{\textcolor{blue}{+0.20}}    & \textit{\textcolor{blue}{+0.44}}    & \textit{\textcolor{blue}{+0.46}}    & \textit{\textcolor{blue}{-0.46 }}       \\ \hline \hline
\rowcolor{black!10}\multicolumn{1}{|l|}{\textbf{\textit{Experiment 2 }}}                     &                  &              &              &              &              \\ \hline
$\pi_{0.5}$          & 0.62            & 0.27         & 0.47         & 0.86         & 0.14         \\ \hline
VA-VLA                            & 0.66             & 0.31         & 0.56         & 0.89         & 0.11         \\ \hline
\multicolumn{1}{|r|}{\textit{\textcolor{blue}{Delta}}}                              & \textit{\textcolor{blue}{+0.04}}            & \textit{\textcolor{blue}{+0.05}}         & \textit{\textcolor{blue}{+0.08 }}        & \textit{\textcolor{blue}{+0.03}}         & \textit{\textcolor{blue}{-0.03}}        \\ \hline \hline
$\pi_{0.5}$  (UC)\footnotemark[1] & 0.58             & 0.28         & 0.41         & 0.81         & 0.19         \\ \hline
VA-VLA (UC)   & 0.64            & 0.32         & 0.48         & 0.87         & 0.13         \\ \hline
\multicolumn{1}{|r|}{\textit{\textcolor{blue}{Delta}} }                             & \textit{\textcolor{blue}{+0.06 }  }         & \textit{\textcolor{blue}{+0.05} }        & \textit{\textcolor{blue}{+0.07}}         & \textit{\textcolor{blue}{+0.06} }        & \textit{\textcolor{blue}{-0.06} }       \\ \hline \hline
$\pi_{0.5}$  (C)       & 0.35            & 0.11         & 0.21         & 0.53         & 0.47         \\ \hline
VA-VLA (C)          & 0.51            & 0.20         & 0.33         & 0.76         & 0.25         \\ \hline
\multicolumn{1}{|r|}{\textit{\textcolor{blue}{Delta}} }                             & \textit{\textcolor{blue}{+0.16} }           & \textit{\textcolor{blue}{+0.09}}         &\textit{ \textcolor{blue}{+0.12 }}        & \textit{\textcolor{blue}{+0.23}}         & \textit{\textcolor{blue}{-0.23}}      \\ \hline
\end{tabular}
\vspace{8pt}
\label{tab:results}
\end{table}
\footnotetext[1]{(C) notes cluttered trials. For experiment 2, (UC) notes uncluttered results for the subset of all tasks that were tested further with clutter.}

\subsection{EC-VLA Results}
The EC-VLA models achieved a higher mean task score on the SO-101 cube picking task than the tuned baseline model, especially in cluttered trials (Table \ref{tab:results}). While the success rate remained constant in the standard trials, the EC-VLA models provided an increase of 20\% success rate in the cluttered case. There was a drastic decrease in complete failure rate, the portion of trials which do not achieve even stage 1 credit. For the baseline model, 50\% of cluttered trials did not touch the target cube, while the EC-VLA model reduced this number to only 6\%. The most common failure mode in this experiment corresponded to correct target identification but unsuccessful grasp execution. We examined the ability of each participant model to recover from such failed grasps, which is provided with other participant-model details in the appendix (Section \ref{sec:ec_recovery}).


\subsection{VA-VLA Results}

Our VA-VLA model generally achieves better performance in the RoboCasa pick-and-place tasks than the tuned baseline $\pi_{0.5}$ model using a full text description. As in the EC-VLA results, this is especially true in the cluttered trials (Table \ref{tab:results}). The most substantial difference between the models is in the complete failure rate of the cluttered setting. While the baseline model fails to even move the target object on 47\% of cluttered trials, the VA-VLA model reduces this to 25\%. These trials seem to be mixed in how much progress they make beyond this point, some only managing to lift the object while others completing the task. We note that numbers here are averaged across tasks, and these trends do not hold for every task individually. However, all tasks saw an increase in performance in cluttered scenes when comparing VA-VLA against the baseline. We provide complete success rate and partial credits scores for each task in the appendix (Tables \ref{tab:taskssuccess} and \ref{tab:taskscompletion}). 


\section{Conclusion and Future Work}

In experiments considering continuous EMG signals and visual annotations as alternatives to fixed language descriptions, we showed that VLAs can be specialized to receive task guidance from non-language modalities that provide task details instead of full text descriptions. We find minor increases in performance for in-distribution scenes with few objects, and strong increases in performance when testing in out-of-distribution scenes with ambiguity or clutter. This indicates that, while language may be important for learning latent representations in model pretraining, it may be less important as a task descriptor for downstream policies.


In addition to data modality, a major change to the baseline in our experiments is that we provide updated conditioning at each state. In experiment 1, we have a human participant supervise policy execution and provide real-time guidance via EMG. In experiment 2, we exploit segmentation abilities from the simulator and provide updated visual annotations that highlight objects of interest. Future work should examine how much guidance is truly needed, and account for possible degradation when the system is deployed with new users or a secondary system to track and segment objects. Also considered should be blending modalities further, using whichever is most convenient for a user or most informative for the policy.




\section*{Acknowledgment}

The authors gratefully acknowledge internal financial support from the Johns
Hopkins Applied Physics Laboratory’s Independent Research and Development (IRAD) Program for
funding portions of this work.


\newpage

\bibliographystyle{ieeetr}
\bibliography{refs}

\newpage
\newpage

\section*{Appendices}

\section{Appendices}

\subsection{Tuning Details and Hyperparameters}

\subsubsection{EC-VLA Tuning Details}
In the EC-VLA experiments, tuning of SmolVLA is conducted with LeRobot \cite{lerobot}. We tune the action expert and state projection while keeping the backbone frozen. We tune the model for 40k updates, using a batch size of 64, and generate a sequence of 50 actions conditioned on a single observation. We use the AdamW \cite{adamw} optimizer with a learning rate of 1e-4 that cosine-decays to 2.5e-6 over 30k steps, a weight decay of 1e-10, and clip gradient norms to 10.0. 

\subsubsection{VA-VLA Tuning Details}
In the VA-VLA experiments, tuning of $\pi_{0.5}$ is conducted with the VLAb port of LeRobot \cite{vlab}. For both models we tune the entire architecture, including the VLM backbone and action expert, and use an action chunk size of 50 conditioned on a single observation. We tune the model for 70k updates on 2xH200s, using a batch size of 64, and apply mild image augmentations to the visual inputs. We use the AdamW optimizer with a learning rate of 2.5e-5 that cosine-decays to 2.5e-6 over 30k steps, a weight decay of 0.01, and clip gradient norms to 1.0. 

\subsection{EC-VLA Data Collection}
\label{sec:ec_data}
To compile demonstrations that include EMG observations, we collect fine-tuning data from three participants under an Institutional Review Board-approved protocol. A participant sits before the SO-101 tabletop setup, and an 8-channel Myo armband EMG (Thalmic Labs, Kitchener, ON, Canada) is placed on the proximal forearm, which is cleaned with isopropyl. During data collection, a separate human operator controls the robot via a leader arm using verbal cues from the participant and without knowledge of which cube was the target. As the participant provides cues, they simultaneously record intent via wrist gestures: flexion (left), extension (right), and relaxation (selection). This results in teleoperated trajectories that have corresponding EMG data to be treated as state information. Each participant provided 105 demonstrations with balanced target locations.

To express the EMG signals as features, the 8-channel EMG signals are rectified and low-pass filtered (at 2Hz) to obtain signal envelopes. Taking the current envelope value for each signal at a given time results in an 8-dimensional state vector that can be concatenated with the existing 6-dimension proprioception vector, creating $q_{EC}$.

\subsection{EC-VLA Participant Models and Recovery Analysis}
\label{sec:ec_recovery}

\begin{figure}[htbp]
    \centering
    \includegraphics[width=0.5\textwidth]{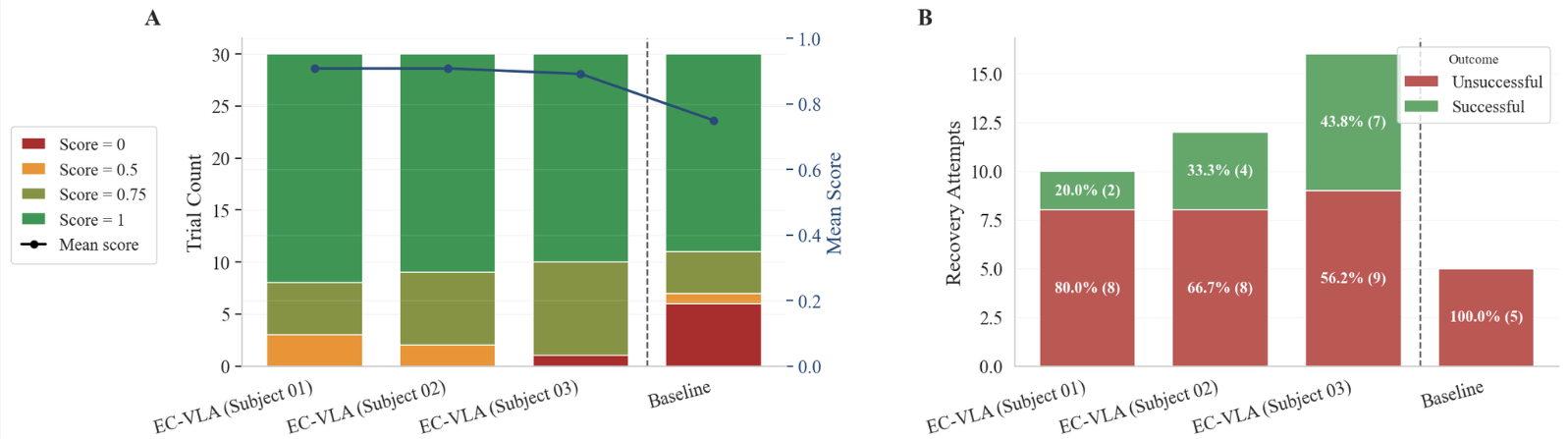}
    \caption{Details of performance for each participant EC-VLA model. \textbf{A:} partial scoring chart for each model compared with the baseline, showing score 0 (Failure, red), score 0.5 (Stage 1, orange), score 0.75 (Stage 2, light green), and score 1.0 (Success, dark green). \textbf{B:} analysis of how often each model attempted to recover from a failed grasp, and whether those attempts were successful.}
    \label{fig:ecappendix}
\end{figure}

We found performance to be very similar among the three participant-specific EC-VLA models in terms of task completion score, as visualized in Figure \ref{fig:ecappendix} A. We also recorded how often each model attempted to recover from failed grasps, and found that participants would attempt corrections more often than the base model. Number of correction attempts were varied among participants, although critically all participants had some successful corrections. The base model, by comparison, never successfully recovered. This analysis is visualized in Figure \ref{fig:ecappendix} B.

\subsection{VA-VLA Task Imagery}

\begin{figure}[htbp]
    \centering
    \includegraphics[width=0.4\textwidth]{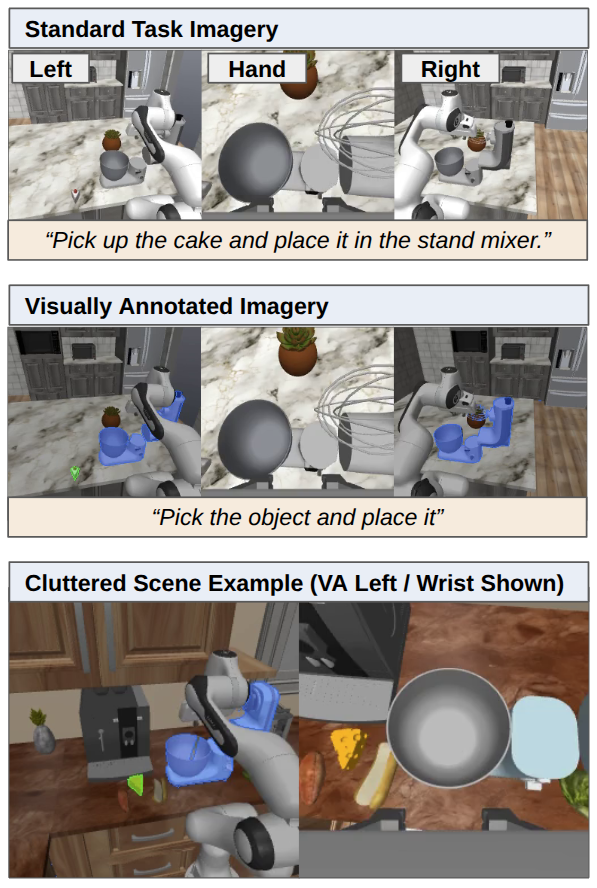}
    \caption{State information used in experiment 2, which always uses a left camera, a wrist camera, and a right camera. In the baseline model, the images are unaltered and paired with a text description (top). In VA-VLA (center), the text is generic and the left and right cameras have visual annotations for the target object (green) and destination (blue). In the cluttered trials (bottom), additional objects are placed in the scene. In practice this uses all three views and is either annotated or not depending on the model.}
    \label{fig:vaimagery}
\end{figure}

Here we provide examples of the image data from experiment two (Figure \ref{fig:vaimagery}).

\subsection{VA-VLA Task-level Results}

Here we provide success rate and task completion score for all 16 tasks examined in RoboCasa (Tables \ref{tab:taskssuccess} and \ref{tab:taskscompletion}).

\newpage

\begin{table*}[]
\centering
\caption{Success rates for all 16 tasks in experiment 2.}
\begin{tabular}{|l|c|c|c|c|}
 \hline
\textbf{Task}                   & \textbf{Lang} & \textbf{Mask} & \textbf{Lang+Clutter} & \textbf{Mask+Clutter} \\ \hline \hline
\textbf{Clutterable Tasks}      &      &      &              &              \\ \hline
Cabinet-to-Counter     & 0.22 & 0.40 & 0.14         & 0.24         \\ \hline
Counter-to-Blender     & 0.20 & 0.18 & 0.02         & 0.02         \\ \hline
Counter-to-Cabinet     & 0.38 & 0.38 & 0.06         & 0.26         \\ \hline
Counter-to-Drawer      & 0.30 & 0.44 & 0.14         & 0.30         \\ \hline
Counter-to-Sink        & 0.30 & 0.28 & 0.18         & 0.16         \\ \hline
Counter-to-StandMixer  & 0.24 & 0.38 & 0.14         & 0.28         \\ \hline
Drawer-to-Counter      & 0.14 & 0.10 & 0.02         & 0.02         \\ \hline
Sink-to-Counter        & 0.42 & 0.42 & 0.20         & 0.30         \\ \hline
\multicolumn{1}{|r|}{\textbf{\textit{Clutterable Mean}}}      & \textbf{\textit{0.275}} & \textbf{\textit{0.323}} & \textbf{\textit{0.113}}         & \textbf{\textit{0.198}}         \\ \hline \hline
\textbf{Non-Clutterable Tasks}  &      &      &              &              \\ \hline
Counter-to-Microwave   & 0.08 & 0.08 &              &              \\ \hline
Counter-to-Oven        & 0.20 & 0.12 &              &              \\ \hline
Counter-to-Stove       & 0.06 & 0.16 &              &              \\ \hline
Counter-to-ToasterOven & 0.46 & 0.38 &              &              \\ \hline
Microwave-to-Counter   & 0.30 & 0.18 &              &              \\ \hline
Stove-to-Counter       & 0.24 & 0.36 &              &              \\ \hline
ToasterOven-to-Counter & 0.44 & 0.54 &              &              \\ \hline
Toaster-to-Counter     & 0.28 & 0.62 &              &              \\ \hline
\multicolumn{1}{|r|}{\textbf{\textit{Non-Clutterable Mean}}}  & \textbf{\textit{0.258}} & \textbf{\textit{0.305}} &              &              \\ \hline \hline
\multicolumn{1}{|r|}{\textbf{\textit{Mean, All Tasks}}}      & \textbf{\textit{0.266}} & \textbf{\textit{0.314}} &              &              \\ \hline
\end{tabular}
\label{tab:taskssuccess}
\end{table*}


\begin{table*}[]
\centering
\caption{Task completion scores for all 16 tasks in experiment 2.}
\begin{tabular}{|l|c|c|c|c|}
 \hline
\textbf{Task}                   & \textbf{Lang} & \textbf{Mask} & \textbf{Lang+Clutter} & \textbf{Mask+Clutter} \\ \hline \hline
\textbf{Clutterable Tasks}      &      &      &              &              \\ \hline
Cabinet-to-Counter     & 0.470 & 0.650 & 0.330        & 0.510        \\ \hline
Counter-to-Blender     & 0.605 & 0.595 & 0.255        & 0.330        \\ \hline
Counter-to-Cabinet     & 0.595 & 0.645 & 0.255        & 0.525        \\ \hline
Counter-to-Drawer      & 0.625 & 0.780 & 0.405        & 0.625        \\ \hline
Counter-to-Sink        & 0.530 & 0.625 & 0.375        & 0.510        \\ \hline
Counter-to-StandMixer  & 0.550 & 0.645 & 0.295        & 0.590        \\ \hline
Drawer-to-Counter      & 0.575 & 0.505 & 0.490        & 0.500        \\ \hline
Sink-to-Counter        & 0.695 & 0.680 & 0.370        & 0.495        \\ \hline
\multicolumn{1}{|r|}{\textbf{\textit{Clutterable Mean}}}      & \textbf{\textit{0.581}} & \textbf{\textit{0.641}} & \textbf{\textit{0.347}}         & \textbf{\textit{0.511}}         \\ \hline \hline
\textbf{Non-Clutterable Tasks}  &      &      &              &              \\ \hline
Counter-to-Microwave   & 0.560 & 0.420 &              &              \\ \hline
Counter-to-Oven        & 0.735 & 0.700 &              &              \\ \hline
Counter-to-Stove       & 0.525 & 0.640 &              &              \\ \hline
Counter-to-ToasterOven & 0.765 & 0.715 &              &              \\ \hline
Microwave-to-Counter   & 0.685 & 0.685 &              &              \\ \hline
Stove-to-Counter       & 0.670 & 0.765 &              &              \\ \hline
ToasterOven-to-Counter & 0.735 & 0.780 &              &              \\ \hline
Toaster-to-Counter     & 0.550 & 0.790 &              &              \\ \hline
\multicolumn{1}{|r|}{\textbf{\textit{Non-Clutterable Mean}}}  & \textbf{\textit{0.653}} & \textbf{\textit{0.687}} &              &              \\ \hline \hline
\multicolumn{1}{|r|}{\textbf{\textit{Mean, All Tasks}}}      & \textbf{\textit{0.617}} & \textbf{\textit{0.664}} &              &              \\ \hline
\end{tabular}
\label{tab:taskscompletion}
\end{table*}


\end{document}